\documentclass[11pt,onecolumn]{IEEEtran}
\usepackage[OT1]{fontenc}

\usepackage{amsfonts}
\IEEEoverridecommandlockouts
\usepackage{bm}
\usepackage{booktabs}
\usepackage{pifont}
\usepackage{floatrow}
\usepackage{times,amsmath,color,amssymb,graphicx,geometry,algorithm}
\usepackage{enumerate}
\usepackage{diagbox}
\usepackage{epstopdf}
\usepackage{multirow}
\usepackage{makecell}
\usepackage{array}
\usepackage{tabularx}
\usepackage{algorithmic}

\usepackage[super,numbers,sort&compress]{natbib}

\usepackage[label font=bf,labelformat=simple]{subfig}
\usepackage[font=small,labelfont=bf,labelsep=period]{caption}

\usepackage[colorlinks,
linkcolor=black,
anchorcolor=black,
citecolor=blue,
filecolor=black,
menucolor=black,
runcolor=black,
urlcolor=black]{hyperref}

\newcommand{\bluecite}[1]{{\hypersetup{citecolor=blue}\cite{#1}}}

\newcommand{\cmark}{\ding{51}}
\newcommand{\nmark}{--}
\newcommand{\datasetroot}{\textit{UAV\_\allowbreak{}full\_\allowbreak{}dataset\_\allowbreak{}release/}}
\newcolumntype{L}[1]{>{\raggedright\arraybackslash}p{#1}}
\newcolumntype{C}[1]{>{\centering\arraybackslash}p{#1}}
\newcolumntype{Y}{>{\raggedright\arraybackslash}X}
\begin{document}

\title{A2A-UAV: A Multimodal Air-to-Air Dataset for UAV Detection and Identification}

\author{
Hongtao Liang\textsuperscript{1,\ensuremath{\dagger}},
Yiyao Wan\textsuperscript{2,\ensuremath{\dagger}},
Kezhi Zhu\textsuperscript{1},
Jiayi Zhang\textsuperscript{1},
Zequn Jiang\textsuperscript{1},
Ziyu Zhou\textsuperscript{1},
Jiahuan Ji\textsuperscript{3},
Fuhui Zhou\textsuperscript{3,*},
and Qihui Wu\textsuperscript{1}\\[0.5em]
\textsuperscript{1}College of Electronic and Information Engineering, Nanjing University of Aeronautics and Astronautics, Nanjing, China\\
\textsuperscript{2}Department of Computer Science and Engineering, The Hong Kong University of Science and Technology, Hong Kong, China\\
\textsuperscript{3}College of Artificial Intelligence, Nanjing University of Aeronautics and Astronautics, Nanjing, China
\thanks{\textsuperscript{\ensuremath{\dagger}}These authors contributed equally to this work.}
\thanks{\textsuperscript{*}Corresponding author: Fuhui Zhou (e-mail: fuhuizhou@ieee.org).}
}

\maketitle

\begin{abstract}
The rapid growth of unmanned aerial vehicles (UAVs) in low-altitude airspace increases the risk of unauthorized intrusions, making reliable detection and identification of non-cooperative UAVs essential for low-altitude airspace safety. 
However, the existing public UAV datasets are mostly acquired from fixed ground-based viewpoints and are limited to imaging or radio-frequency (RF) observations, which restricts the sensing coverage and deployment flexibility. 
To address this gap, we present A2A-UAV, a multimodal air-to-air dataset for UAV detection and identification. 
Specifically, 15 target UAVs of 13 commercial models were recorded at three outdoor sites under daytime and nighttime conditions by using a UAV-mounted platform equipped with an RGB--thermal gimbal camera, a panoramic camera, an RF spectrum analyzer, and a millimeter-wave radar. 
The dataset contains 105,294 time-aligned samples of RGB, panoramic, thermal, and RF time--frequency images and radar point clouds. 
Moreover, all 209,539 bounding boxes of the predominantly small UAV targets were manually annotated and cross-checked. 
Therefore, A2A-UAV provides a valuable resource for air-to-air UAV detection, identification, and multimodal fusion in low-altitude airspace.

\end{abstract}

\newpage
\IEEEpeerreviewmaketitle

\section*{Background \& Summary}
The rapid development of unmanned aerial vehicles (UAVs) has enabled their wide deployment in logistics, infrastructure inspection, environmental monitoring, and emergency response\bluecite{civil_drones_nature}. 
However, the low cost and accessibility of commercial UAVs also facilitate unauthorized operations, such as flights in restricted airspace and unauthorized surveillance of sensitive facilities, posing threats to public safety and critical infrastructure\bluecite{dutantiuav}. 
The effective regulation and countermeasure of such UAVs rely on the timely perception of non-cooperative UAV targets, which do not broadcast their identity or position. 
Therefore, accurate detection and identification of non-cooperative UAVs are prerequisites for airspace monitoring, early warning, and subsequent countermeasures.

Various sensing modalities have been investigated for the detection and identification of non-cooperative UAVs. 
Specifically, electro-optical cameras capture the appearance and shape of the target, and thermal infrared cameras record the thermal contrast between the target and the background{\hypersetup{citecolor=blue}\textcolor{black}{\cite{antiuav,huang_uavcb}}}. 
Radar sensors measure target range and radial velocity through the time delay and Doppler shift of reflected electromagnetic waves, independently of illumination conditions. 
Moreover, passive radio-frequency (RF) sensors capture communication and control signals between UAVs and their remote controllers, providing complementary model-related information through spectral occupancy, frequency-hopping patterns, and modulation characteristics\bluecite{hanjung2026}. 

Recent advances in deep learning have promoted data-driven methods for UAV detection and identification{\hypersetup{citecolor=blue}\textcolor{black}{\cite{huang_uavcb,yu_rf_openset}}}. 
The development and evaluation of these methods rely on large-scale sensor observations with reliable annotations. 
Therefore, representative public datasets with standardized evaluation protocols are essential for the development, benchmarking, and reproducible comparison of these methods{\hypersetup{citecolor=blue}\textcolor{black}{\cite{hanjung2026,zhang_tdtiv}}}.

Table~\ref{tab:dataset_gap} summarizes representative public UAV datasets in terms of the acquisition setting, the sensing modalities, and the availability of category labels. 
The Drone-vs-Bird\bluecite{dronevsbird} and DUT Anti-UAV\bluecite{dutantiuav} datasets provide ground-to-air RGB observations for UAV detection and tracking. 
Moreover, the Anti-UAV\bluecite{antiuav} benchmark and the multi-sensor drone detection dataset\bluecite{dronedetection} provide both visible and thermal infrared observations. 
Furthermore, the DroneRF\bluecite{dronerf}, CardRF\bluecite{cardrf}, and RFUAV\bluecite{rfuav} datasets provide RF recordings for UAV detection and identification, which were acquired by ground-based receivers or under controlled conditions. 
However, these datasets rely on sensors at fixed ground locations, which constrains the available viewpoints and exposes the target observations to occlusion by buildings and vegetation.

To overcome these limitations, UAV-mounted sensing provides a flexible platform for the detection and identification of non-cooperative UAVs{\hypersetup{citecolor=blue}\textcolor{black}{\cite{liang_synergetic_empowerment}}}. 
Specifically, the airborne observer can adjust its position and viewing direction to maintain line-of-sight observations of the targets when ground-level observations are obstructed.
\textcolor{black}{However, air-to-air observation differs fundamentally from ground-based observation. The simultaneous motion of the sensing platform and the target UAV causes continuous variations in viewing direction, target scale, and background, which alternates among the sky, terrain, buildings, and vegetation. Models trained on ground-to-air data, which are acquired from static viewpoints against predominantly sky backgrounds, are therefore not directly transferable to airborne observation, and dedicated air-to-air datasets are required.}
Several datasets have been developed to investigate UAV perception in this air-to-air setting.
The Det-Fly\bluecite{detfly} dataset provides RGB images for air-to-air micro-UAV detection, while the MOT-FLY\bluecite{motfly} dataset provides visual sequences for multi-UAV tracking. 
Recently, the MMFW-UAV\bluecite{mmfwuav} dataset introduced visible and thermal infrared imagery for fixed-wing UAV detection.

Despite these advances, the existing air-to-air datasets are restricted to imaging modalities, which limits the joint analysis of heterogeneous sensing information. 
Recent studies have shown that the complementary information from heterogeneous sensors can improve the robustness of UAV perception compared with single-modality approaches\bluecite{uav_multimodal_tmc,hanjung2026}.
\textcolor{black}{However, the existing RF and radar UAV datasets\bluecite{dronerf,cardrf,rfuav,hanjung2026} were acquired by ground-based sensors or under controlled conditions, and they generally do not provide visual observations of the same targets. As a result, the RF emissions and radar returns of UAVs observed from an airborne platform, as well as their association with the visual appearance of the same targets, cannot be investigated by using the existing public datasets.}
Moreover, the existing air-to-air datasets rely on cameras with a limited field of view, which cover only a small part of the surrounding airspace.
In contrast, panoramic cameras capture the full surrounding airspace of the observer in a single frame, which enables the omnidirectional surveillance of non-cooperative UAVs without prior knowledge of the target direction.
Therefore, the lack of panoramic \textcolor{black}{observations and airborne non-imaging observations with visual counterparts} limits the development and evaluation of UAV detection and identification methods that exploit omnidirectional and heterogeneous sensing information.

\begin{table}[t]
\begingroup
\centering
\caption{Comparison of representative UAV datasets. A check mark and ``--'' indicate documented and undocumented features, respectively. Pano., TIR, RF, and ID denote panoramic imagery, thermal infrared imagery, radio-frequency data, and category-level identification based on multiple UAV labels, respectively.}
\label{tab:dataset_gap}
\footnotesize
\renewcommand{\baselinestretch}{1.0}\selectfont
\setlength{\tabcolsep}{2.5pt}
\renewcommand{\arraystretch}{1.15}
\begin{tabular*}{\textwidth}{@{\extracolsep{\fill}}L{2.45cm}L{2.40cm}C{0.95cm}C{1.00cm}C{0.90cm}C{0.80cm}C{1.10cm}C{0.75cm}C{1.45cm}@{}}
\toprule
Dataset & \makecell[l]{Acquisition\\setting} & RGB & Pano. & TIR & RF & Radar & ID & \makecell{Release\\date} \\
\midrule
\textcolor{black}{Drone-vs-Bird\cite{dronevsbird}} & \textcolor{black}{Ground-to-air} & \textcolor{black}{\cmark} & \textcolor{black}{\nmark} & \textcolor{black}{\nmark} & \textcolor{black}{\nmark} & \textcolor{black}{\nmark} & \textcolor{black}{\nmark} & \textcolor{black}{Sep. 2019} \\
DUT Anti-UAV\cite{dutantiuav} & Ground-to-air & \cmark & \nmark & \nmark & \nmark & \nmark & \nmark & May 2020 \\
Anti-UAV\cite{antiuav} & Ground-to-air & \cmark & \nmark & \cmark & \nmark & \nmark & \nmark & Feb. 2020 \\
Drone-det.\cite{dronedetection} & Ground-to-air & \cmark & \nmark & \cmark & \nmark & \nmark & \nmark & May 2020 \\
\midrule
\textcolor{black}{DroneRF\cite{dronerf}} & \textcolor{black}{Controlled RF} & \textcolor{black}{\nmark} & \textcolor{black}{\nmark} & \textcolor{black}{\nmark} & \textcolor{black}{\cmark} & \textcolor{black}{\nmark} & \textcolor{black}{\cmark} & \textcolor{black}{Oct. 2019} \\
\textcolor{black}{CardRF\cite{cardrf}} & \textcolor{black}{Ground-based RF} & \textcolor{black}{\nmark} & \textcolor{black}{\nmark} & \textcolor{black}{\nmark} & \textcolor{black}{\cmark} & \textcolor{black}{\nmark} & \textcolor{black}{\cmark} & \textcolor{black}{2022} \\
RFUAV\cite{rfuav} & \textcolor{black}{Controlled RF} & \nmark & \nmark & \nmark & \cmark & \nmark & \cmark & Mar. 2025 \\
\midrule
Det-Fly\cite{detfly} & Air-to-air & \cmark & \nmark & \nmark & \nmark & \nmark & \nmark & Dec. 2020 \\
MOT-FLY\cite{motfly} & Air-to-air & \cmark & \nmark & \nmark & \nmark & \nmark & \nmark & Feb. 2023 \\
MMFW-UAV\cite{mmfwuav} & Air-to-air & \cmark & \nmark & \cmark & \nmark & \nmark & \nmark & May 2024 \\
\midrule
\textbf{A2A-UAV} & \textbf{Air-to-air} & \textbf{\cmark} & \textbf{\cmark} & \textbf{\cmark} & \textbf{\cmark} & \textbf{\cmark} & \textbf{\cmark} & \textbf{Pending} \\
\bottomrule
\end{tabular*}
\endgroup
\end{table}

To address these limitations, we constructed A2A-UAV, a multimodal dataset for air-to-air UAV detection and identification. 
The data were acquired by a UAV-mounted platform equipped with an RGB--thermal gimbal camera, a panoramic camera, an RF spectrum analyzer, and a millimeter-wave radar. 
The platform observed 15 target UAVs of 13 commercial models, including three separately labeled DJI Mini 4 Pro units, in playground, park, and garden scenes under daytime and nighttime conditions. 
To the best of our knowledge, A2A-UAV is the first air-to-air UAV dataset that provides \textcolor{black}{time-aligned RGB, panoramic, thermal, RF, and radar} observations of the same targets with category-level labels. 
The main contributions of this dataset are summarized as follows.\nopagebreak[4]

\begin{itemize}

\item \textcolor{black}{\textbf{Five-modality air-to-air observations.}
A2A-UAV provides time-aligned RGB, panoramic, thermal infrared, RF time--frequency, and millimeter-wave radar observations of UAV targets acquired from an airborne platform. Specifically, the panoramic images cover the full surrounding airspace, the RF time--frequency images capture the model-related transmission characteristics, and the radar point clouds provide the spatial and radial-velocity measurements of the detected returns.}

\item \textcolor{black}{\textbf{Real-flight data with high-quality annotations.}
The dataset comprises 105,294 samples acquired in three scenes under daytime and nighttime conditions. All 209,539 bounding boxes of the predominantly small UAV targets were manually annotated without automatic pre-annotation and subsequently cross-checked.}

\item \textcolor{black}{\textbf{Diverse commercial UAV models.}
The 15 UAV labels of A2A-UAV cover 13 commercial models of different airframe designs, equipped with five transmission-system variants spanning several generations from OcuSync~2.0 to DJI O4+. 
This diversity provides varied visual appearances and RF signatures, which supports the development and evaluation of UAV identification methods across models and transmission systems.}

\item \textcolor{black}{\textbf{Support for flexible model development.}
A2A-UAV links the time-aligned observations, annotations, and metadata of each sample through a common sample identifier, which supports both single-modality and multimodal methods for UAV detection and identification. 
The predefined recording-level training, validation, and test splits further provide a common basis for reproducible method comparison.}
\end{itemize}

\begin{figure}[t]
\centering
\includegraphics[page=2,width=\textwidth]{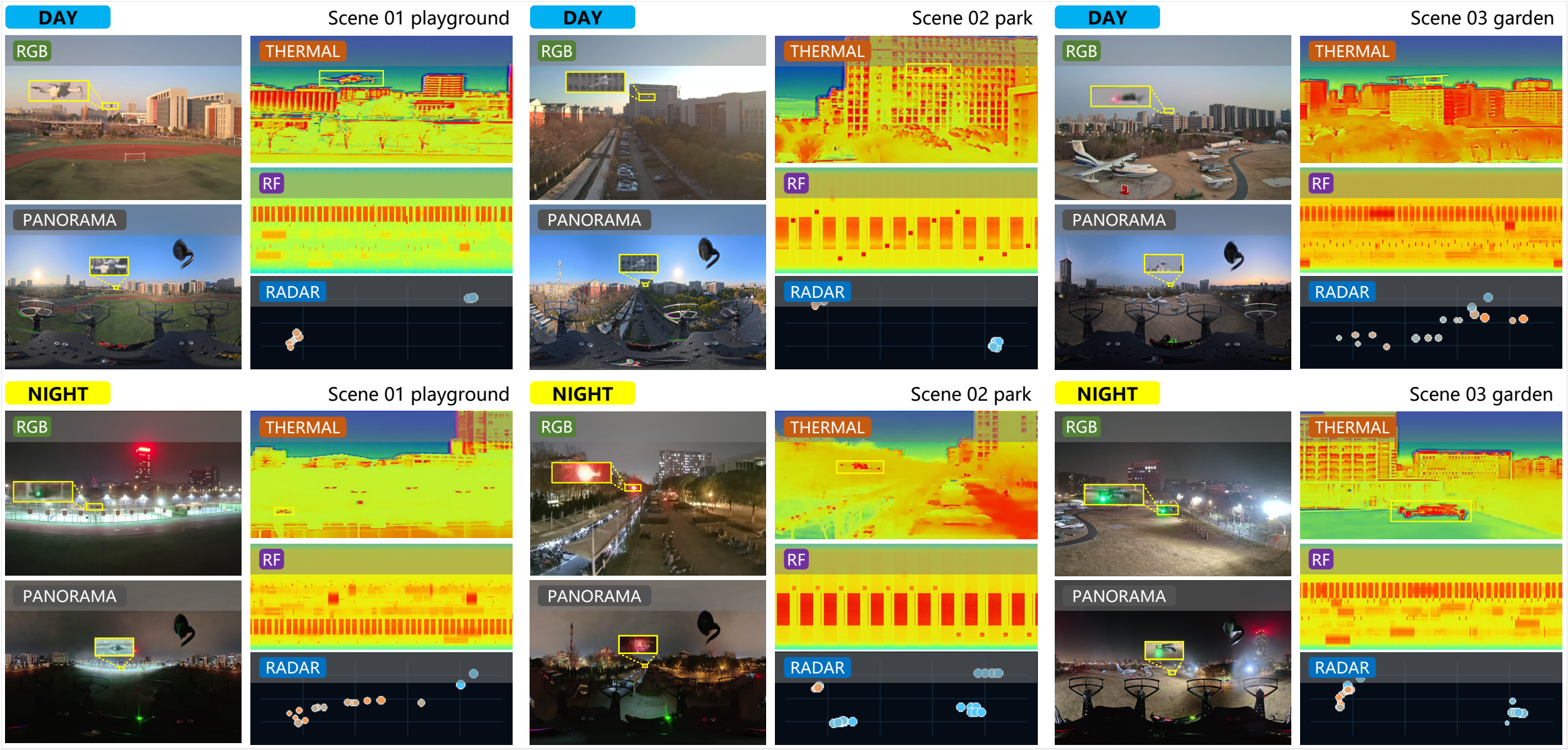}
\caption{Representative multimodal observations from A2A-UAV across different scenes and illumination conditions.}
\label{fig:dataset_examples}
\end{figure}

\textcolor{black}{The acquisition platform and procedure, multimodal sample construction, annotation and quality control, data organization, and technical validation of A2A-UAV are detailed in the following sections.}
Fig.~\ref{fig:dataset_examples} presents representative observations across the three scenes and both illumination conditions.

\section*{Methods}

\begin{figure}[t]
\centering
\includegraphics[width=\textwidth]{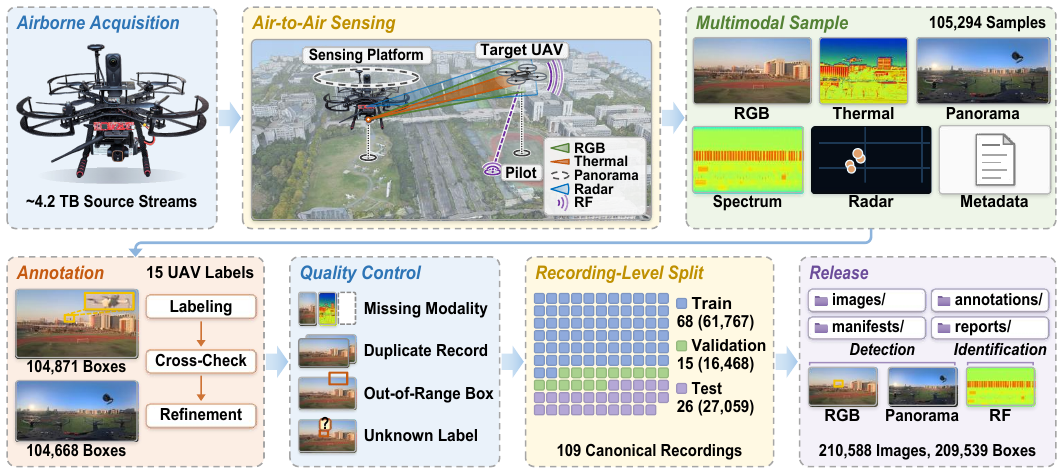}
\caption{Data construction and release workflow of A2A-UAV.}
\label{fig:processing_pipeline_current}
\end{figure}

The A2A-UAV was built through five stages, namely, the data acquisition, multimodal sample construction, manual annotation, quality control, and recording-level partitioning, as summarized in Fig.~\ref{fig:processing_pipeline_current}. 
The following subsections describe the sensing platform, acquisition sites, and data processing procedures.

\subsection*{Data Collection Platform}
\begin{figure}[t]
\centering
\includegraphics[width=\textwidth]{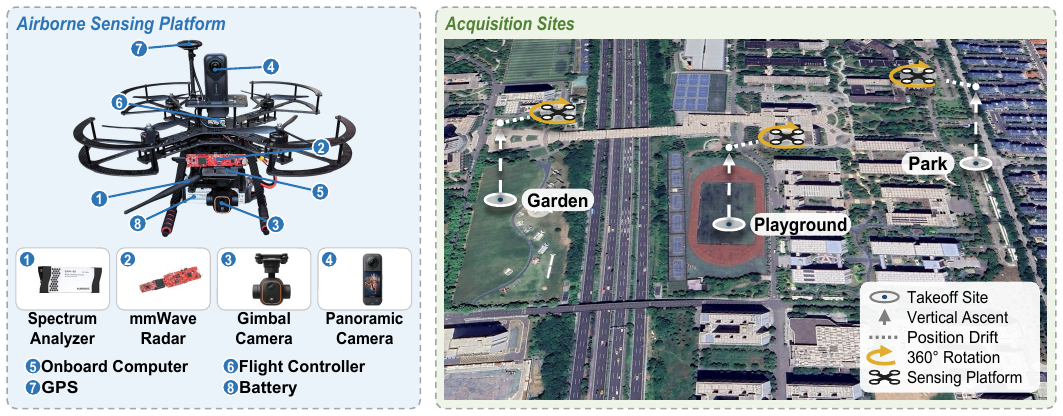}
\caption{Airborne sensing platform and acquisition sites of A2A-UAV.}
\label{fig:sensing_platform_current}
\end{figure}

The airborne data acquisition platform integrates four sensing devices to acquire five data modalities, as shown in the left panel of Fig.~\ref{fig:sensing_platform_current}. 
Specifically, the RGB--thermal gimbal camera captures the RGB and thermal infrared images, and the panoramic camera records the $360^\circ$ videos. 
Moreover, the RF spectrum analyzer records the RF signals between the target UAV and its remote controller, and the millimeter-wave radar generates the point clouds of the surrounding airspace. 
The principal settings of these devices are summarized in Table~\ref{tab:platform_specifications} and described below.

\begin{enumerate}[1)]
\color{black}
\item The RF spectrum analyzer is a HAROGIC NXM-60 networked real-time analyzer\cite{harogic_nxm60}.
\textcolor{black}{The analyzer was configured with a center frequency of 2.42~GHz, an IQ sampling rate of 125~MSa/s, with the in-phase and quadrature components stored as 16-bit integers.}

\item \textcolor{black}{The millimeter-wave radar unit uses a Texas Instruments IWR1843AOPEVM module with an integrated frequency-modulated continuous-wave (FMCW) sensor\cite{ti_iwr1843aop}. During data acquisition, the radar operated with three transmit and four receive channels. Each chirp started at 77~GHz and covered an effective sweep bandwidth of approximately 3.32~GHz. Moreover, the frame period was set to 50~ms.}

\item The RGB--thermal gimbal camera is a Skydroid C12, which combines a 5-megapixel RGB camera and a thermal infrared camera on a three-axis stabilized mount\cite{skydroid_c12}.
Specifically, the RGB camera has an F/2.0 aperture, a field of view of $100^\circ \times 52^\circ$, and a video resolution of $1280 \times 720$ pixels.
Moreover, the thermal camera has a resolution of $384 \times 288$ pixels, a spectral range of 8--14~$\mu\mathrm{m}$, a field of view of $24.8^\circ \times 18.7^\circ$, and a frame \textcolor{black}{rate of 25~Hz}.

\normalcolor
\item The panoramic camera is an Insta360 X5, which employs two 1/1.28-inch sensors with F/2.0 lenses and provides $360^\circ$ angular coverage\cite{insta360_x5}.
During data acquisition, the camera recorded panoramic video at $5760 \times 2880$ pixels and \textcolor{black}{30}~fps.
\end{enumerate}

\begin{table}[t]
\begingroup
\centering
\footnotesize
\renewcommand{\baselinestretch}{1.0}\selectfont
\setlength{\tabcolsep}{4.5pt}
\renewcommand{\arraystretch}{1.14}
\begin{tabularx}{\textwidth}{@{}L{3.05cm}L{4.20cm}Y@{}}
\toprule
Device & Parameter & Specification \\
\midrule
\multirow{3}{3.05cm}{HAROGIC NXM-60}
& Center frequency & 2.42~GHz \\
& IQ sampling rate & 125~MSa/s \\
& \textcolor{black}{IQ component bit depth} & \textcolor{black}{16 bits} \\
\midrule
\multirow{4}{3.05cm}{Texas Instruments IWR1843AOPEVM}
& \textcolor{black}{Chirp start frequency} & \textcolor{black}{77~GHz} \\
& \textcolor{black}{ADC-sampled bandwidth} & \textcolor{black}{3.32~GHz} \\
& \textcolor{black}{Frame period} & \textcolor{black}{50~ms} \\
& Antenna configuration & Three transmit and four receive antennas \\
\midrule
\multirow{9}{3.05cm}{Skydroid C12}
& Mount & Three-axis stabilized gimbal \\
& RGB sensor resolution & 5 megapixels \\
& RGB aperture & F/2.0 \\
& RGB field of view & $100^\circ$ horizontal and $52^\circ$ vertical \\
& RGB video resolution & $1280 \times 720$ pixels \\
& Thermal resolution & $384 \times 288$ pixels \\
& Thermal spectral range & 8--14~$\mu\mathrm{m}$ \\
& Thermal field of view & $24.8^\circ \times 18.7^\circ$ \\
& Thermal frame rate & \textcolor{black}{25~Hz} \\
\midrule
\multirow{5}{3.05cm}{Insta360 X5}
& Sensor configuration & Two 1/1.28-inch sensors \\
& Aperture & F/2.0 \\
& Angular coverage & $360^\circ$ \\
& Video resolution & $5760 \times 2880$ pixels \\
& Frame rate & \textcolor{black}{30}~fps \\
\bottomrule
\end{tabularx}
\caption{Airborne sensing platform specifications.}
\label{tab:platform_specifications}
\endgroup
\end{table}

In addition to the sensing devices, the airborne platform carries an onboard computer for data logging, a flight-control system, a GPS receiver, and a battery module.
The data were collected at three outdoor sites on the campus of Nanjing University of Aeronautics and Astronautics in Nanjing, China.
Specifically, the sites are the playground (31.937223$^\circ$ N, 118.794636$^\circ$ E), the park (31.938202$^\circ$ N, 118.798412$^\circ$ E), and the garden (31.936469$^\circ$ N, 118.792271$^\circ$ E). 
The locations of the three sites are shown in the right panel of Fig.~\ref{fig:sensing_platform_current}.

The target set comprised 15 individual UAVs representing 13 commercial models, including three separately labeled DJI Mini 4 Pro units, as summarized in Table~\ref{tab:target_uav_specs}. 
Specifically, the dimensions, maximum horizontal speeds, and transmission-system designations listed in the table were obtained from the corresponding manufacturer specifications~\cite{dji_aircraft_specs,dji_mavic4_manual,dji_neo_controls}. 
During each recording session, the airborne sensing platform took off first and established a stable observation position.
The target UAV then entered the monitored airspace, and both aircraft were manually piloted with a safe separation maintained between them.
The operator adjusted the position and heading of the sensing platform and the orientation of the camera gimbal to keep the target within the effective sensing region during repeated flight passes. 
Moreover, the relative altitude and bearing between the two UAVs were varied across the passes, introducing changes in viewing direction, target scale, and background.
The acquisition procedure was repeated at the three sites under daytime and nighttime conditions to capture variations in background and illumination.
The operators of both the sensing platform and the target UAVs held valid civil unmanned aircraft pilot licenses issued by the Civil Aviation Administration of China (CAAC)\cite{caac_uav_regulations}.

\begin{table}[t]
\begingroup
\centering
\small
\renewcommand{\baselinestretch}{1.0}\selectfont
\setlength{\tabcolsep}{5pt}
\renewcommand{\arraystretch}{1.20}
\begin{tabularx}{\textwidth}{@{}L{3.4cm}C{2.8cm}>{\centering\arraybackslash}X C{2.2cm}@{}}
\toprule
\makecell[l]{UAV model} & \makecell{Transmission\\system} & \makecell{Dimensions\\L $\times$ W $\times$ H (mm)} & \makecell{Max. speed\\(m/s)} \\
\midrule
DJI Avata & DJI O3+ & $180 \times 180 \times 80$ & 14 \\
DJI Air 2S & DJI O3 & $180 \times 97 \times 77$ & 19 \\
DJI Air 3 & DJI O4 & $207 \times 100.5 \times 91.1$ & 21 \\
DJI Air 3S & DJI O4 & $214.19 \times 100.63 \times 89.17$ & 21 \\
DJI Avata 2 & DJI O4 & $185 \times 212 \times 64$ & 16 \\
DJI Flip & DJI O4 & $136 \times 62 \times 165$ & 12 \\
DJI Mavic 2 Pro & OcuSync 2.0 & $214 \times 91 \times 84$ & 20 \\
DJI Mavic 4 Pro & DJI O4+ & $257.6 \times 124.8 \times 103.4$ & 25 \\
DJI Mini 3 Pro & DJI O3 & $145 \times 90 \times 62$ & 16 \\
DJI Mini 4 Pro & DJI O4 & $148 \times 94 \times 64$ & 16 \\
DJI Mini 5 Pro & DJI O4+ & $157 \times 95 \times 68$ & 18 \\
DJI Neo & DJI O4 & $130 \times 157 \times 48.5$ & 8 \\
DJI Neo 2 & DJI O4 & $147 \times 171 \times 41$ & 12 \\
\bottomrule
\end{tabularx}
\caption{Target UAV models, transmission systems, and manufacturer specifications.}
\label{tab:target_uav_specs}
\endgroup
\end{table}

\subsection*{Data Extraction and Multimodal Sample Construction}
The raw sensor recordings were processed into sample-level multimodal data products following the workflow illustrated in Fig.~\ref{fig:processing_pipeline_current}. 
During flight operations, the onboard computer recorded RGB, thermal, and radar streams in Robot Operating System (ROS) bag files. 
The RGB and thermal frames were extracted and saved as PNG images at their recorded resolutions of $1280 \times 720$ and $384 \times 288$ pixels, respectively. 
RGB frames were stored as 24-bit PNG images (8 bits per channel), while thermal frames retained the bit depth and channel configuration of the recorded ROS image messages. Panoramic frames were extracted from the Insta360 X5 videos and saved as 24-bit PNG images at a resolution of $5760 \times 2880$ pixels. 
Moreover, the radar messages were converted into point-cloud records, while the raw radio-frequency (RF) in-phase and quadrature (IQ) recordings were retained for the subsequent signal processing.

The recorded RF IQ data were converted into time--frequency images by using the short-time Fourier transform (STFT)\textcolor{black}{, and each RF acquisition recorded an IQ capture of 0.125~s. To make full use of each capture, three partially overlapping 0.1-s segments were extracted from its beginning, center, and end, and the central time of each segment was used as its acquisition timestamp. 
Therefore, each RF time--frequency image is associated with the actual acquisition time of the IQ samples that it represents.} For each segment, the mean complex value was subtracted to remove the direct-current component. A two-sided STFT was then computed by using a 1024-sample Hamming window, a 512-sample overlap, and a 1024-point fast Fourier transform (FFT).
The resulting spectrum was shifted to center the zero-baseband-frequency bin, corresponding to the receiver center frequency of 2.42~GHz. The squared magnitudes of the STFT coefficients were converted to a logarithmic power scale and visualized by using the jet color map.
The resulting time--frequency images were saved as $6000 \times 3600$-pixel RGBA PNG files.

\textcolor{black}{RGB, thermal, and panoramic frames were extracted by using reference timestamps generated at 4~Hz on a common session-level time axis. Because small and distant UAV targets produce sparse radar returns, all radar frames within $\pm120$~ms of the reference timestamp were assigned to the sample to increase the probability of capturing the target, with each frame saved as an individual CSV file. This 240-ms window is shorter than the 250-ms reference interval, ensuring that each radar frame belongs to at most one sample. Samples without any radar frame within the window were marked as radar-unavailable in the metadata. }

\textcolor{black}{
The RF analyzer sampled the IQ data at 125~MSa/s with 16-bit in-phase and quadrature components, corresponding to a raw data rate of approximately 500~MB/s. To limit the volume of RF data collected during each flight session, RF signals were acquired as discrete 0.125-s IQ captures. 
Accordingly, instead of frame-level alignment with the imaging streams, each sample was assigned the RF time--frequency image whose acquisition timestamp is nearest to its reference timestamp within the same recording. 
This association is consistent with the role of the RF modality in A2A-UAV, which provides model-related transmission signatures for the UAV identification. 
Specifically, each sample was assigned the RF time--frequency image with the nearest acquisition timestamp within the same recording. 
Each recording contains a single target UAV, and the hopping pattern, occupied bandwidth, and frame structure of the RF emissions are determined by the transmission system of the target rather than by its instantaneous position relative to the sensing platform. 
Therefore, the nearest RF time--frequency image provides a representative transmission signature of the target in the sample. 
Moreover, the samples that share the same RF time--frequency image belong to the same recording and are assigned to the same partition, which prevents RF information leakage across partitions.
}

\subsection*{Data Annotation Procedures}
The RGB and panoramic images were annotated separately with bounding boxes. 
All annotations were performed manually by using Label Studio~\cite{labelstudio}, without automatic pre-annotation, frame interpolation, or tracking-based propagation. 
Specifically, each image was annotated individually following a common guideline that required bounding boxes to tightly enclose the visible extent of the UAV, including the fuselage, arms, and propellers. 
For partially occluded or truncated targets, the bounding boxes were drawn according to the visible portions of the UAV. 
Moreover, each bounding box was assigned a UAV label based on the flight logs, which identified the target UAV in each recording session.

\subsection*{Data Quality Control}
\textcolor{black}{Following the labeling stage in Fig.~\ref{fig:processing_pipeline_current}, all bounding boxes underwent a manual cross-check and refinement, and the missing, duplicate, inaccurate, and incorrectly labeled annotations were corrected. 
The exported records were then validated for missing modalities, duplicate records, out-of-range boxes, and unknown labels. 
Specifically, the samples with a missing or unreadable image modality and the duplicate records were removed. 
Moreover, the bounding boxes beyond the image boundaries were corrected, and the labels outside the 15 predefined UAV labels were reassigned. 
Furthermore, the samples with more than one bounding box in either image or without a bounding box in both images were excluded. 
The 1,049 image records without a bounding box were retained, because the target could leave the field of view of one camera while remaining visible in the other. 
Finally, a privacy audit was conducted on the RGB images by using automatic face and license-plate detectors, and the confirmed faces were blurred in the released images. }

\subsection*{Dataset Partitioning}

The A2A-UAV dataset was partitioned at the recording level after annotation and quality control. Each canonical recording identifier represents a continuous recording of a single target UAV acquired in one scene under one illumination condition. All samples generated from the same recording group were assigned to the same partition, and their associated modality files and annotations were retained with the corresponding samples. The training, validation, and test sets contain 68, 15, and 26 recording groups, respectively. 
Table~\ref{tab:dataset_summary_current} summarizes \textcolor{black}{the sample and annotation counts for each partition, together with the total radar-sample and radar-frame counts}. This recording-level design keeps temporally adjacent samples from the same recording within a single partition, which reduces the risk of information leakage. All 15 UAV labels are represented in each partition, providing label coverage for subsequent evaluation on the validation and test sets.

\begin{table}[!t]
\centering
\caption{Composition of the full A2A-UAV dataset.}
\label{tab:dataset_summary_current}
\begin{tabular}{lrrrr}
\toprule
Record type & Train & Validation & Test & Total \\
\midrule
Samples & 61,767 & 16,468 & 27,059 & 105,294 \\
COCO RGB/panorama image records & 123,534 & 32,936 & 54,118 & 210,588 \\
COCO RGB/panorama boxes & 122,762 & 32,934 & 53,843 & 209,539 \\
\textcolor{black}{Samples with radar data} & \textcolor{black}{61,737} & \textcolor{black}{16,468} & \textcolor{black}{27,058} & \textcolor{black}{105,263} \\
\textcolor{black}{Radar frames} & \textcolor{black}{296,794} & \textcolor{black}{79,327} & \textcolor{black}{129,552} & \textcolor{black}{505,673} \\
\bottomrule
\end{tabular}
\end{table}

\section*{Data Records}

\begingroup
\hypersetup{urlcolor=black,linkcolor=black}
\color{black}
A 200-sample preview subset of A2A-UAV is hosted on Google Drive at \url{https://drive.google.com/file/d/1oCtjoF5vfH3MsPDNQEOQ84MvRywHSUso/view}.
An identical mirror is provided on Baidu Netdisk at \url{https://pan.baidu.com/s/1_P6F3z18gCyuDagdedcM2Q?pwd=A2AD} with the extraction code A2AD.
The full dataset comprises 271 independently extractable \texttt{.tar.gz} archives with a total size of approximately 2.50~TB, and extracting all archives into the same directory restores the structure shown in Fig.~\ref{fig:repository_structure_current}.
The full dataset will be publicly released upon acceptance of this paper, and the preview subset follows the same directory structure and file formats.
\endgroup
The full A2A-UAV dataset includes multimodal data, metadata, bounding-box annotations, and predefined data splits.
This section details the repository organization, file-naming conventions, and characteristics of the full dataset.

\subsection*{\textcolor{black}{Repository Structure}}

\begin{figure}[!t]
\centering
\includegraphics[height=0.68\textheight,width=\textwidth,keepaspectratio]{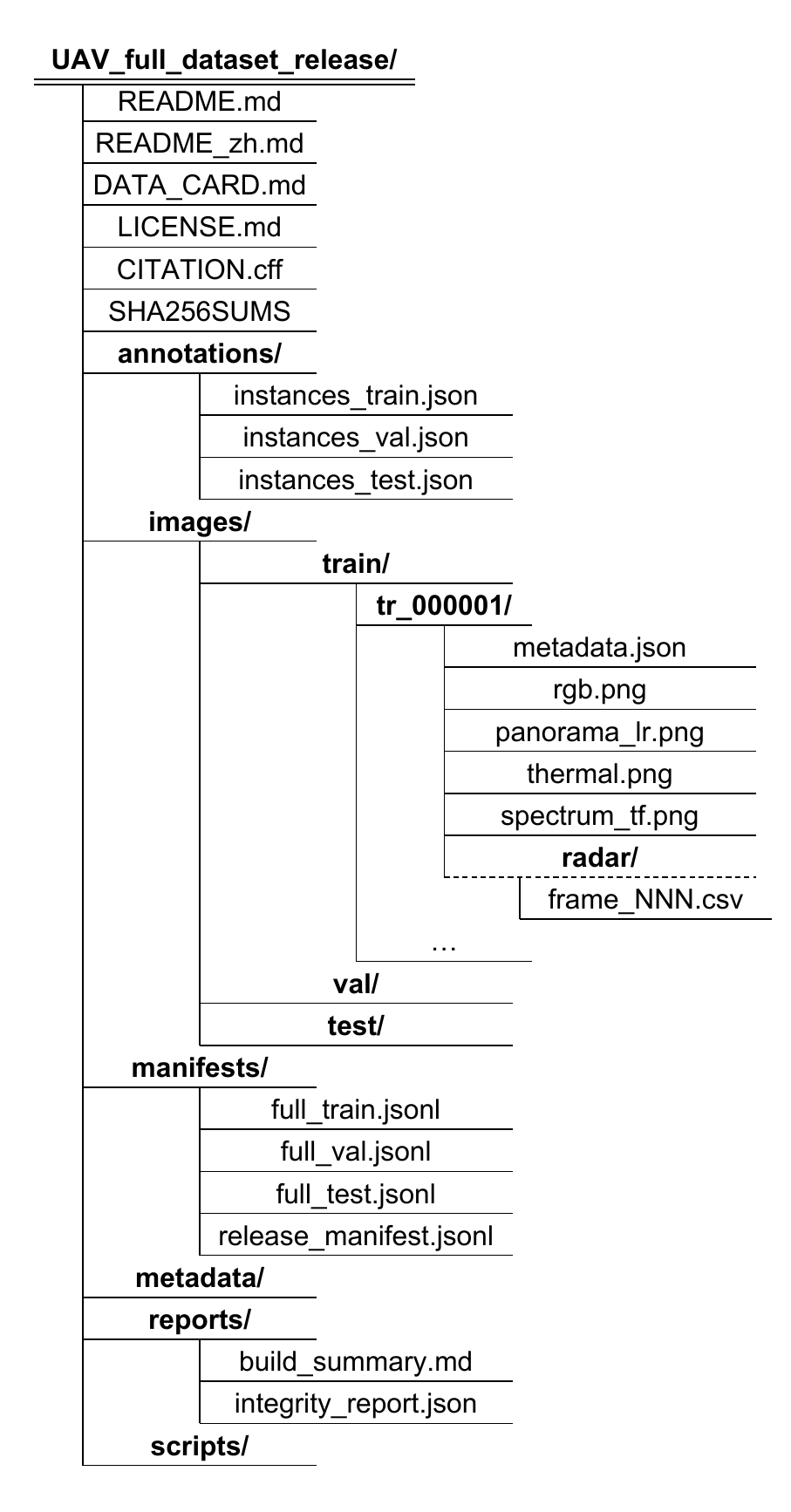}
\caption{\textcolor{black}{Repository structure of A2A-UAV.}}
\label{fig:repository_structure_current}
\end{figure}

Fig.~\ref{fig:repository_structure_current} illustrates the organization of the A2A-UAV release, with \datasetroot{} as the root directory. 
Specifically, the release is organized into six main subdirectories, namely \textit{images/}, \textit{annotations/}, \textit{manifests/}, \textit{reports/}, \textcolor{black}{\textit{metadata/}}, \textcolor{black}{and \textit{scripts/}}. 
The \textit{images/} directory stores the multimodal products and sample metadata in the predefined \textit{train/}, \textit{val/}, and \textit{test/} splits, with one directory for each sample. The \textit{annotations/} directory contains the corresponding COCO-style bounding-box annotations. 
The \textit{manifests/} directory provides sample-level JSON Lines (JSONL) records describing split membership, modality paths, and annotation metadata, while the \textit{reports/} directory contains the release summary and integrity report. 
\textcolor{black}{Additional metadata and supporting scripts are organized under \textit{metadata/} and \textit{scripts/}, respectively.}
Moreover, the root directory also includes documentation files describing the dataset, licensing information, citation metadata, and file checksums.

\subsection*{\textcolor{black}{File Naming}}

Sample directories follow the pattern \textit{images/\textless{}split\textgreater{}/\textless{}sample\_id\textgreater{}/}. The split names \textit{train}, \textit{val}, and \textit{test} denote the training, validation, and test sets. Each sample identifier combines the corresponding prefix \textit{tr}, \textit{va}, or \textit{te} with a six-digit anonymized sample number. For example, \textit{images/train/tr\_000001/} denotes the directory for the training sample \textit{tr\_000001}.
Within each sample directory, \textit{rgb.png}, \textit{panorama\_lr.png}, \textit{thermal.png}, and \textit{spectrum\_tf.png} store the RGB, panoramic, thermal, and RF time--frequency images, respectively. The file \textit{metadata.json} stores the associated sample-level metadata. \textcolor{black}{When radar observations are available, each radar frame is stored as a separate CSV file under \textit{radar/}, following the pattern \textit{radar/frame\_NNN.csv}, where \textit{NNN} is a zero-padded three-digit frame index.} The same sample identifier is recorded in the metadata, JSONL split manifests, and RGB and panoramic COCO image records. 
Moreover, this shared identifier links the modality files and bounding-box annotations to the corresponding sample. 

\subsection*{Data Properties}

Each sample directory contains the RGB, panoramic, thermal, and RF time--frequency images and a metadata file. 
Specifically, the radar point cloud is available for 105,263 samples, and the 31 retained samples had no radar frame within the temporal window and were marked as radar-unavailable in the metadata. 
The 105,263 samples contain 505,673 radar frames in total, and each sample with radar observations is associated with one to six frames. 
Table~\ref{tab:modalities_current} summarizes the format and availability of each released product. 
The \textit{metadata.json} file records the sample identifier, the split assignment, the scene and illumination labels, the modality availability, and the RGB and panoramic annotations. 
Moreover, all image products are stored as lossless PNG files and the RF time--frequency images are rendered by using the jet color map. 
The horizontal axis spans the 0.1-s segment, the vertical axis spans 2.3575 to 2.4825~GHz, and the color represents the relative logarithmic power within each image. 
Furthermore, each radar-frame CSV file stores one point per row with the fields \texttt{x}, \texttt{y}, \texttt{z}, \texttt{intensity}, and \texttt{velocity}. 
The coordinates are given in meters in the radar sensor frame, and the velocity denotes the radial velocity in meters per second.

The bounding boxes are provided in two formats. 
The \textit{metadata.json} files and the JSONL manifests record each box as $[x_{\min}, y_{\min}, x_{\max}, y_{\max}]$ in pixel coordinates, with the origin at the top-left corner of the image. 
The COCO-style annotation files follow the standard COCO format $[x_{\min}, y_{\min}, w, h]$. 
Moreover, each box is assigned one of the 15 UAV labels listed in Table~\ref{tab:label_counts_current}, and each COCO-style image record includes a \texttt{modality} field that distinguishes the RGB images from the panoramic images. 

\begin{table}[t]
\begingroup
\centering
\small
\renewcommand{\baselinestretch}{1.0}\selectfont
\setlength{\tabcolsep}{5pt}
\renewcommand{\arraystretch}{1.2}
\renewcommand{\tabularxcolumn}[1]{m{#1}}
\caption{Released multimodal products, properties, and availability.}
\label{tab:modalities_current}
\begin{tabularx}{\textwidth}{@{}>{\raggedright\arraybackslash}m{3.1cm}>{\raggedright\arraybackslash}m{5.5cm}Y r@{}}
\toprule
\textbf{Product} & \textbf{File and properties} & \textbf{Role} & \makecell[r]{\textbf{Available}\\\textbf{samples}} \\
\midrule
RGB image & \textit{rgb.png}\newline $1280 \times 720$ pixels & Visual target appearance & 105,294 \\
\addlinespace[3pt]
Panoramic image & \textit{panorama\_lr.png}\newline $5760 \times 2880$ pixels & Wide-field visual context & 105,294 \\
\addlinespace[3pt]
Thermal image & \textit{thermal.png}\newline $384 \times 288$ pixels & Thermal target observations & 105,294 \\
\addlinespace[3pt]
RF time--frequency\newline image & \textit{spectrum\_tf.png}\newline $6000 \times 3600$ pixels & RF activity representation & 105,294 \\
\addlinespace[3pt]
Radar point cloud & \textit{radar/frame\_NNN.csv}\newline One CSV file per radar frame\newline \texttt{x}, \texttt{y}, \texttt{z}, \texttt{intensity}, \texttt{velocity} & Spatial and motion information & 105,263 \\
\addlinespace[3pt]
Metadata & \textit{metadata.json}\newline Sample fields and RGB/panorama boxes & Sample association and labels & 105,294 \\
\bottomrule
\end{tabularx}
\endgroup
\end{table}

\subsection*{Dataset Analysis}

The A2A-UAV dataset contains 105,294 samples with 15 UAV labels. 
Specifically, the playground, park, and garden scenes contribute 30,113, 43,502, and 31,679 samples, respectively, and the daytime and nighttime conditions contribute 52,162 and 53,132 samples, respectively. 
Each sample provides one RGB image and one panoramic image for annotation, which yields 210,588 image records and 209,539 bounding boxes. 
Moreover, the number of boxes per label ranges from 9,098 to 20,135, as listed in Table~\ref{tab:label_counts_current}. 

\begin{figure}[t]
\centering
\includegraphics[width=\textwidth]{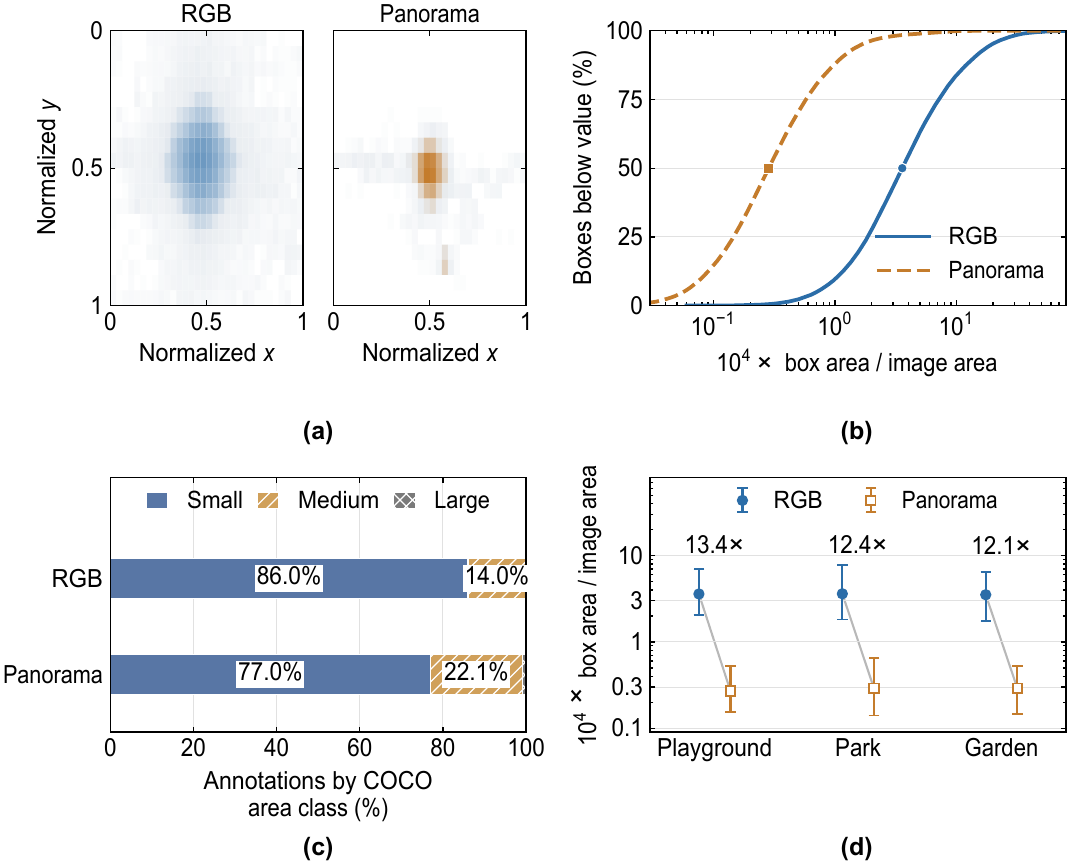}
\caption{Target positions and scales in A2A-UAV. \textbf{(a)} Distributions of the normalized box centers. \textbf{(b)} Cumulative distributions of the normalized box area, with markers indicating the medians. \textbf{(c)} Proportions of the COCO area classes. \textbf{(d)} Median normalized box areas in the three scenes, with error bars indicating the interquartile ranges  and values giving the RGB-to-panoramic ratios.}
\label{fig:annotation_scale_position_current}
\end{figure}

Fig.~\ref{fig:annotation_scale_position_current} characterizes the target positions and scales. 
The RGB box centers concentrate around the image center because the gimbal was oriented to keep the target within the field of view. 
The panoramic box centers concentrate near the horizontal center line, which corresponds to the near-horizontal lines of sight between the two UAVs. 
Moreover, the median ratio of the box area to the image area is $3.58\times10^{-4}$ for the 104,871 RGB boxes and $2.84\times10^{-5}$ for the 104,668 panoramic boxes. 
Furthermore, 86.0\% of the RGB boxes and 77.0\% of the panoramic boxes fall within the COCO small-object range below $32^2$ pixels\cite{coco}. 
The panoramic targets have a median area of approximately 470 pixels, which exceeds the approximately 330 pixels of the RGB targets because of the higher panoramic resolution. 
The RGB-to-panoramic ratio of the median normalized area remains between 12.1 and 13.4 across the three scenes.

\begin{figure}[t]
\centering
\includegraphics[width=\textwidth]{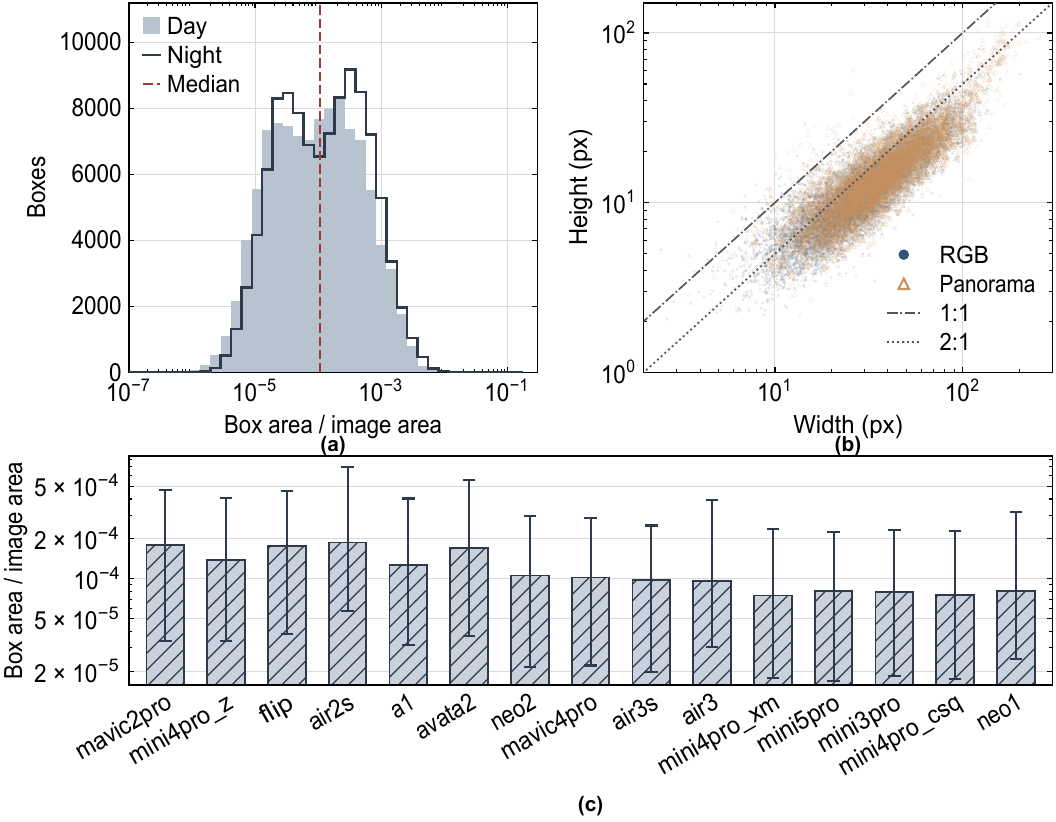}
\caption{Target-scale distributions in A2A-UAV. \textbf{(a)} Daytime and nighttime histograms of the normalized box area, with the dashed line indicating the median. \textbf{(b)} Width--height scatter distributions of RGB and panoramic bounding boxes. \textbf{(c)} Median normalized box area of each UAV label, with error bars indicating the interquartile ranges.}
\label{fig:target_scale_distribution_current}
\end{figure}

Fig.~\ref{fig:target_scale_distribution_current} further compares the target scales across illumination conditions, box shapes, and UAV labels. 
The daytime and nighttime scale distributions overlap substantially. 
Moreover, most bounding boxes have width-to-height ratios greater than $2{:}1$, as shown by the concentration of points below the corresponding reference line. 
Furthermore, the median area ratio of every UAV label remains on the order of $10^{-4}$. 
Therefore, the small-target characteristic holds across illumination conditions, imaging modalities, and UAV labels.

\begin{figure}[t]
\centering
\includegraphics[width=\textwidth]{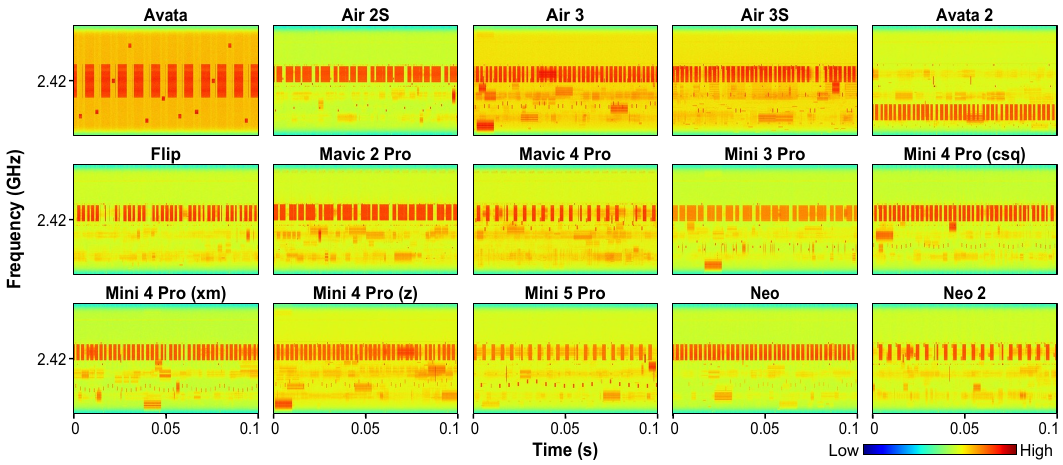}
\caption{Representative RF time--frequency images for the 15 UAV labels in A2A-UAV.}
\label{fig:rf_spectrograms_current}
\end{figure}

Fig.~\ref{fig:rf_spectrograms_current} presents representative RF time--frequency images for the 15 UAV labels. 
The transmission systems of the target UAVs maintain a bidirectional link between the UAV and the remote controller, and the two link directions exhibit distinct time--frequency signatures. 
Specifically, the uplink command-and-control signal from the remote controller appears as short bursts of a few megahertz that hop across the 2.4~GHz band. 
Moreover, the downlink signal from the UAV carries the video and telemetry streams and appears as wideband OFDM blocks with a bandwidth of 10 to 40~MHz. 
Furthermore, the burst duration, the hopping interval, the occupied bandwidth, and the frame periodicity vary across UAV models. 
Therefore, different UAV models exhibit distinguishable time--frequency patterns, and the three DJI Mini 4 Pro units exhibit closely similar patterns. 
These model-dependent signatures provide the discriminative information for RF-based UAV identification.

\begin{table}[t]
\begingroup
\centering
\small
\renewcommand{\baselinestretch}{1.0}\selectfont
\setlength{\tabcolsep}{4pt}
\renewcommand{\arraystretch}{1.2}
\caption{RGB and panoramic bounding-box counts by UAV model and label.}
\label{tab:label_counts_current}
\begin{tabularx}{\textwidth}{@{}Y L{2.75cm}>{\raggedleft\arraybackslash}p{1.3cm}@{\hspace{1.8em}}Y L{2.75cm}>{\raggedleft\arraybackslash}p{1.3cm}@{}}
\toprule
\textbf{UAV model} & \textbf{Label} & \textbf{Boxes} & \textbf{UAV model} & \textbf{Label} & \textbf{Boxes} \\
\midrule
DJI Avata & \texttt{a1} & 18,626 & DJI Mavic 4 Pro & \texttt{mavic4pro} & 15,619 \\
DJI Air 2S & \texttt{air2s} & 13,538 & DJI Mini 3 Pro & \texttt{mini3pro} & 11,157 \\
DJI Air 3 & \texttt{air3} & 17,024 & DJI Mini 4 Pro & \texttt{mini4pro\_csq} & 14,188 \\
DJI Air 3S & \texttt{air3s} & 13,246 & DJI Mini 4 Pro & \texttt{mini4pro\_xm} & 14,311 \\
DJI Avata 2 & \texttt{avata2} & 20,135 & DJI Mini 4 Pro & \texttt{mini4pro\_z} & 11,843 \\
DJI Flip & \texttt{flip} & 16,374 & DJI Mini 5 Pro & \texttt{mini5pro} & 10,667 \\
DJI Mavic 2 Pro & \texttt{mavic2pro} & 10,845 & DJI Neo & \texttt{neo1} & 12,868 \\
DJI Neo 2 & \texttt{neo2} & 9,098 & & & \\
\bottomrule
\end{tabularx}
\endgroup
\end{table}

\section*{Technical Validation}

UAV detection and identification are two fundamental tasks for low-altitude airspace monitoring. 
Specifically, UAV detection localizes the target UAV in the image coordinate system, and UAV identification determines the category of the target UAV. 
Therefore, A2A-UAV is designed to support the detection and identification of UAVs from an air-to-air perspective. 

\subsection*{Model Implementation Details}

The detection benchmarks were implemented by using MMDetection\cite{mmdetection}, and all 15 UAV labels were mapped to a single foreground category. 
Specifically, Faster R-CNN\cite{fasterrcnn}, Dynamic R-CNN\cite{dynamicrcnn}, Cascade R-CNN\cite{cascadercnn}, Sparse R-CNN\cite{sparsercnn}, DINO\cite{dino}, DDQ\cite{ddq}, and FCOS-KL\cite{kldet} were evaluated on both the RGB and panoramic images. 
All detectors used ResNet-50 backbones. 
Moreover, each RGB image was resized to fit within $1706 \times 960$ pixels while preserving its aspect ratio, and each panoramic image was divided into 18 tiles of $1280 \times 1280$ pixels with a stride of 1024 pixels. 
The tile predictions were mapped back to the original image coordinates and merged by the non-maximum suppression with an IoU threshold of 0.55. 
The experiments were conducted by using PyTorch~2.1.0 and MMDetection~3.3.0 on NVIDIA GeForce RTX~4090 GPUs. 
The detection performance was evaluated by using the COCO metrics, where AP averages the precision over IoU thresholds from 0.50 to 0.95, and AP$_S$, AP$_M$, and AP$_L$ correspond to the boxes with areas below $32^2$, between $32^2$ and $96^2$, and above $96^2$ pixels, respectively. 

The identification benchmark used the RF time--frequency images as the input and covered all 15 UAV labels. 
Specifically, the $6000 \times 3600$-pixel RGBA spectrograms were converted into three-channel images of $1333 \times 800$ pixels. 
Six identification networks were evaluated, namely ResNet-18 and ResNet-50\cite{resnet}, EfficientNet-V2-S\cite{efficientnetv2}, ConvNeXt-Tiny\cite{convnext}, Swin-T\cite{swin}, and MobileNet-V3-Large\cite{mobilenetv3}. 
The checkpoint with the highest validation macro-F1 score was selected for testing.

\subsection*{UAV Detection Results}

The left half of Table~\ref{tab:detection_bench_current} presents the UAV detection results on the panoramic test images. 
Specifically, Faster R-CNN achieved the highest AP of 0.1983, AP50 of 0.5010, AP$_S$ of 0.1660, and AP$_M$ of 0.3264. 
Moreover, Cascade R-CNN achieved the highest AP75 of 0.1203, and Dynamic R-CNN achieved the highest AP$_L$ of 0.0514. 
Furthermore, the AP$_S$ of the panoramic baselines ranged from 0.0507 to 0.1660. 
The panoramic targets occupy a median of $2.84\times10^{-5}$ of the image area, and the equirectangular projection further distorts the target appearance. 
Therefore, the small target size and the projection distortion make panoramic UAV detection the more challenging of the two protocols. 
The low AP$_L$ results from the scarcity of large panoramic targets, which account for less than 1\% of the panoramic boxes.

\begin{table}[t]
\centering
\caption{UAV detection results on panoramic (left) and RGB (right) test images. Bold values indicate the best result for each metric within each modality.}
\label{tab:detection_bench_current}
\begingroup
\small
\setlength{\tabcolsep}{1.8pt}
\renewcommand{\arraystretch}{1.08}
\begin{tabular*}{\textwidth}{@{\extracolsep{\fill}}lrrrrrr@{\hspace{10pt}}rrrrrr@{}}
\toprule
\multirow{2}{*}{Model} & \multicolumn{6}{c}{Panorama} & \multicolumn{6}{c}{RGB} \\
\cmidrule(lr){2-7}\cmidrule(lr){8-13}
 & AP & AP50 & AP75 & AP$_S$ & AP$_M$ & AP$_L$ & AP & AP50 & AP75 & AP$_S$ & AP$_M$ & AP$_L$ \\
\midrule
Faster R-CNN\cite{fasterrcnn} & \textbf{0.1983} & \textbf{0.5010} & 0.1183 & \textbf{0.1660} & \textbf{0.3264} & 0.0489 & 0.2810 & \textbf{0.6850} & 0.1750 & 0.2650 & 0.3890 & 0.1500 \\
Dynamic R-CNN\cite{dynamicrcnn} & 0.1868 & 0.4822 & 0.1051 & 0.1564 & 0.3128 & \textbf{0.0514} & 0.2810 & \textbf{0.6850} & 0.1730 & 0.2640 & 0.3920 & \textbf{0.4840} \\
Cascade R-CNN\cite{cascadercnn} & 0.1947 & 0.4793 & \textbf{0.1203} & 0.1653 & 0.3134 & 0.0510 & \textbf{0.2860} & 0.6790 & \textbf{0.1860} & \textbf{0.2700} & \textbf{0.3990} & 0.1510 \\
Sparse R-CNN\cite{sparsercnn} & 0.1711 & 0.4851 & 0.0717 & 0.1549 & 0.2514 & 0.0158 & 0.2420 & 0.6370 & 0.1320 & 0.2340 & 0.3070 & 0.0500 \\
DINO\cite{dino} & 0.1504 & 0.4412 & 0.0568 & 0.1382 & 0.2152 & 0.0005 & 0.2460 & 0.6440 & 0.1300 & 0.2330 & 0.3420 & 0.1510 \\
DDQ\cite{ddq} & 0.1604 & 0.4492 & 0.0733 & 0.1441 & 0.2211 & 0.0087 & 0.2440 & 0.6530 & 0.1190 & 0.2300 & 0.3600 & 0.2020 \\
FCOS-KL\cite{kldet} & 0.0499 & 0.1340 & 0.0243 & 0.0507 & 0.0487 & 0.0000 & 0.2510 & 0.6550 & 0.1360 & 0.2370 & 0.3460 & 0.0000 \\
\bottomrule
\end{tabular*}
\endgroup
\end{table}

The right half of Table~\ref{tab:detection_bench_current} presents the results on the RGB test images.  
Specifically, Cascade R-CNN achieved the highest AP of 0.2860, AP75 of 0.1860, AP$_S$ of 0.2700, and AP$_M$ of 0.3990. 
Moreover, Faster R-CNN and Dynamic R-CNN achieved the same highest AP50 of 0.6850, and Dynamic R-CNN achieved the highest AP$_L$ of 0.4840. 
Compared with the panoramic protocol, the best AP and the best AP$_S$ on the RGB images were higher by 0.0877 and 0.1040, respectively. 
The higher RGB performance results from the larger relative target size and the absence of projection distortion. 

Across both protocols, the highest values of all metrics were achieved by the two-stage detectors, namely Faster R-CNN, Cascade R-CNN, and Dynamic R-CNN, which refine region proposals on high-resolution feature maps and thereby preserve the features of small UAV targets. 
Moreover, AP75 was substantially lower than AP50, with the best AP75 of 0.1203 and 0.1860 compared with the best AP50 of 0.5010 and 0.6850 on the panoramic and RGB images, respectively. 
For small targets, an offset of a few pixels causes a large reduction in IoU. 
Specifically, an offset of 3 pixels in both directions reduces the IoU of an $18\times18$-pixel box, which corresponds to the median RGB target size, to 0.53. 
Therefore, future air-to-air UAV detectors should preserve the fine-grained features of small targets, improve the localization precision, and account for the projection distortion of panoramic images.

\begin{figure}[t]
\centering
\includegraphics[width=\textwidth]{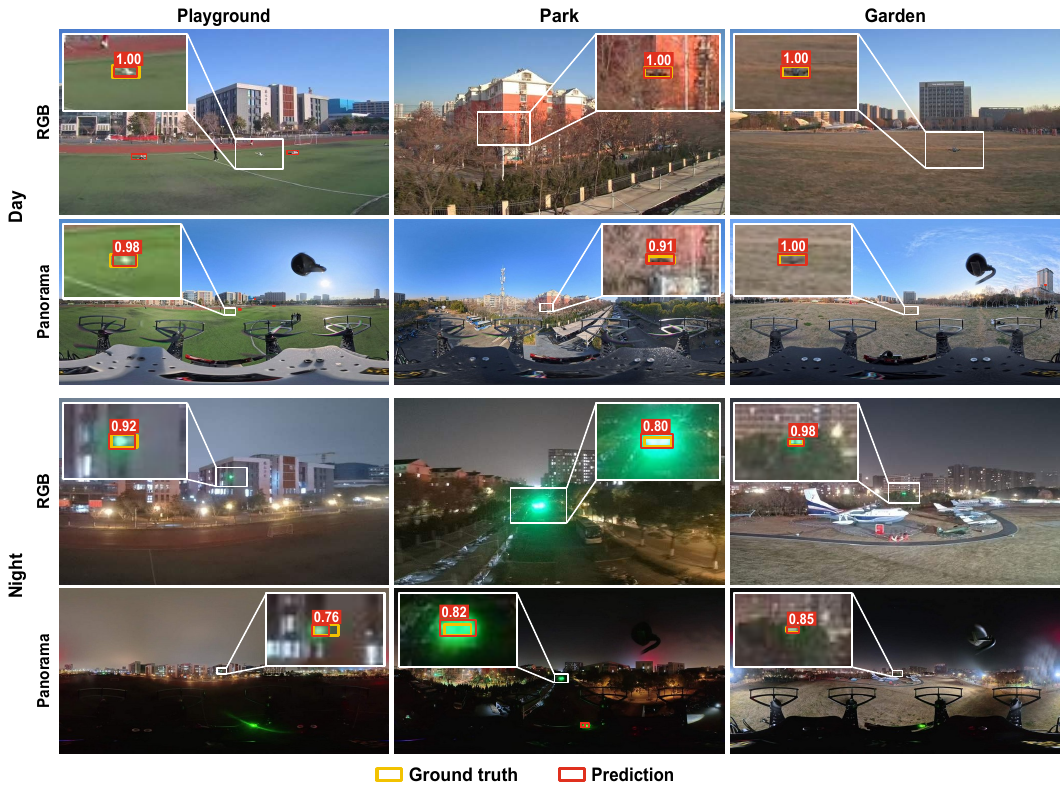}
\caption{Faster R-CNN detections on RGB and panoramic validation images.}
\label{fig:detection_visualization_current}
\end{figure}

Fig.~\ref{fig:detection_visualization_current} further presents the Faster R-CNN detections on paired RGB and panoramic validation images across the three scenes under daytime and nighttime conditions, with all predictions of confidence scores of at least 0.5 displayed. 
In the daytime images, the UAV targets were detected with confidence scores between 0.91 and 1.00. 
Moreover, the confidence scores decreased to between 0.76 and 0.98 in the nighttime images, where the UAV appears mainly as a light spot without a visible contour. 
Furthermore, false positives occurred on background regions that resemble the appearance of distant UAV targets. 
Therefore, the discrimination between small UAV targets and similar background objects remains a key challenge for air-to-air UAV detection.

\subsection*{UAV Identification Results}

Table~\ref{tab:rf_identification_current} demonstrates the RF-based identification results of the six identification networks on the test images over the 15 UAV labels. 
The accuracy measures the proportion of correctly classified samples, the balanced accuracy averages the recall over the 15 labels, and the macro-F1 averages the F1 score over the 15 labels with equal weight. 
EfficientNet-V2-S achieved the highest accuracy of 69.32\%, balanced accuracy of 78.46\%, and macro-F1 of 74.17\%. 
Moreover, MobileNet-V3-Large achieved the second-highest values of all three metrics, and both lightweight networks outperformed the deeper ResNet-50. 

\begin{figure}[t]
\centering
\includegraphics[width=0.85\textwidth]{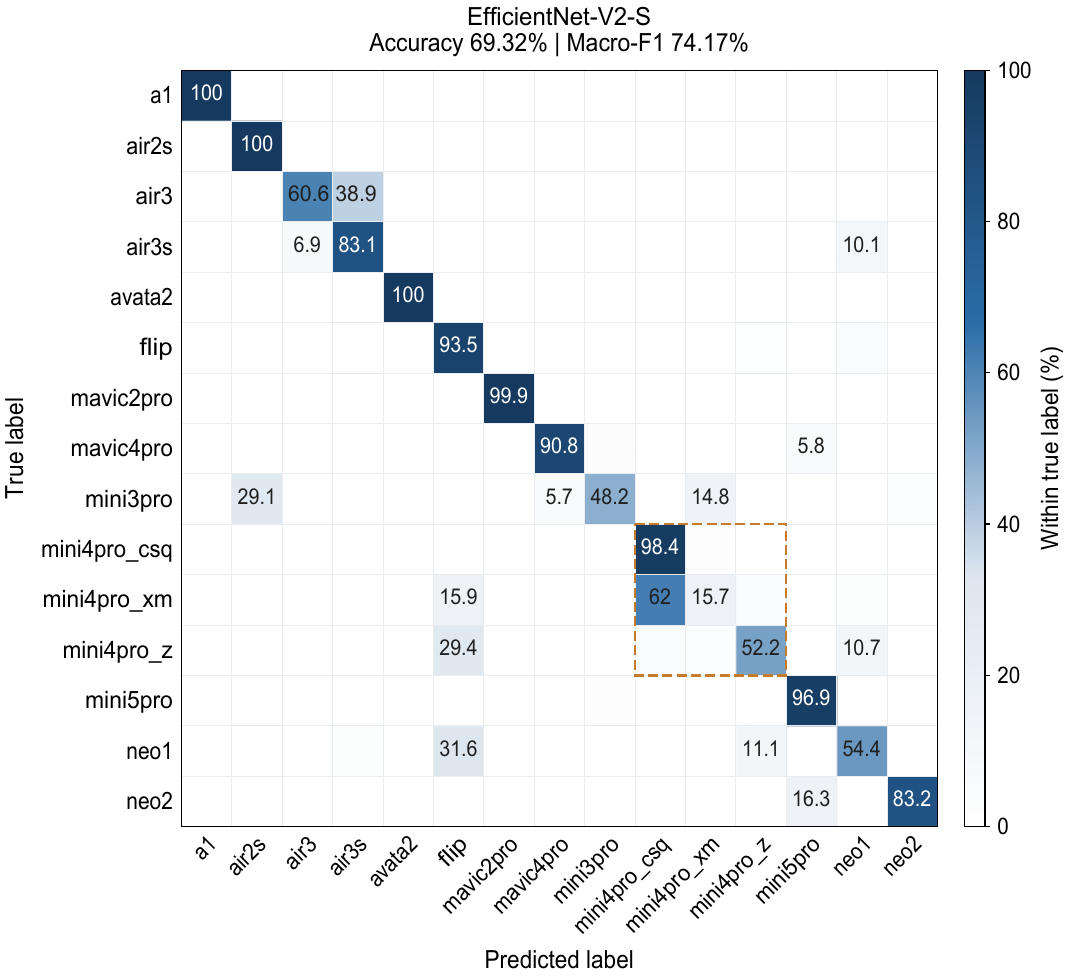}
\caption{{Row-normalized test confusion matrix (\%) of EfficientNet-V2-S for the 15 UAV labels, with the dashed square marking the three Mini 4 Pro aircraft.}}
\label{fig:rf_identification_confusion_current}
\end{figure}

Fig.~\ref{fig:rf_identification_confusion_current} presents the row-normalized confusion matrix of EfficientNet-V2-S on the test images, where the diagonal entries represent the recall of each label. 
The recall reached 100\% for \texttt{a1}, \texttt{air2s}, and \texttt{avata2}, and decreased to 15.70\%, 48.20\%, and 52.24\% for \texttt{mini4pro\_xm}, \texttt{mini3pro}, and \texttt{mini4pro\_z}, respectively. 
Specifically, 62.02\% of the \texttt{mini4pro\_xm} samples were classified as \texttt{mini4pro\_csq}, and the confusion among the three DJI Mini 4 Pro units accounted for 32.79\% of all test errors. 
The three units share an identical transmission protocol, and their differences are limited to subtle hardware-induced characteristics. 
Moreover, 38.92\% of the \texttt{air3} samples were classified as \texttt{air3s}, and the two models share the same O4 transmission system. 
Therefore, future RF-based UAV identification methods should exploit fine-grained hardware characteristics beyond the protocol-level time--frequency patterns to distinguish the UAV models and individual units that share the same transmission system.

\begin{table}[t]
\begingroup
\centering
\caption{RF-based UAV identification results on the test images. All metrics are reported in percent.}
\label{tab:rf_identification_current}
\begin{tabular}{lrrr}
\toprule
Model & Accuracy & Balanced accuracy & Macro-F1 \\
\midrule
ResNet-18 & 54.86 & 64.37 & 58.35 \\
ResNet-50 & 52.25 & 59.07 & 56.38 \\
EfficientNet-V2-S & \textbf{69.32} & \textbf{78.46} & \textbf{74.17} \\
ConvNeXt-Tiny & 63.73 & 74.57 & 69.71 \\
Swin-T & 60.95 & 70.75 & 68.72 \\
MobileNet-V3-Large & 65.74 & 75.49 & 71.57 \\
\bottomrule
\end{tabular}
\endgroup
\end{table}

\section*{Usage Notes}

\textcolor{black}{The dataset provides synchronized RGB, panoramic, thermal, RF time--frequency, and radar observations of 13 commercial UAV models in air-to-air scenarios, together with bounding-box annotations in both the per-sample JSON format and the COCO format. 
Moreover, the three acquisition scenes and the daytime and nighttime conditions cover a wide range of backgrounds and illumination levels. 
Therefore, A2A-UAV supports several research and engineering applications as follows.}

\subsection*{Air-to-Air UAV Detection and Identification}

The RGB and panoramic images and the corresponding bounding boxes support the training and evaluation of air-to-air UAV detectors. 
Specifically, the panoramic images cover the full surrounding airspace of the carrier UAV, and the scene and illumination labels enable the evaluation of detection robustness across backgrounds and between daytime and nighttime conditions. 
Moreover, the RF time--frequency images capture the control and video links of 13 UAV models across several generations of DJI transmission systems, which supports the development of RF-based UAV identification methods. 
Therefore, A2A-UAV is applicable to airborne UAV surveillance, reconnaissance, and collision avoidance.

\subsection*{Multimodal Fusion for UAV Perception}

All modality files of a sample share a common sample identifier. 
Therefore, the visual, thermal, RF, and radar observations of the same target can be paired directly for feature-level or decision-level fusion. 
Specifically, the thermal images and radar point clouds provide observations that are less sensitive to illumination than the visible images, and the RF time--frequency images provide category information that is independent of the target appearance. 
Moreover, the radar availability flag in the metadata enables the selection of the 105,263 samples with radar observations for fusion experiments that require all modalities.

\subsection*{Potential Applications with Secondary Development}

A2A-UAV currently provides bounding boxes for the RGB and panoramic images. 
However, the dataset can be extended to further tasks through secondary annotation. 
Specifically, the bounding boxes can be transferred to the thermal images to support thermal UAV detection, and the radar points can be associated with the visual targets to support radar-based UAV localization. 
Moreover, the separation of UAV labels between training and evaluation supports open-set and few-shot UAV identification. 
These extensions further enhance the application value of A2A-UAV in the airborne UAV perception.

\section*{Data Availability}

\begingroup
\hypersetup{urlcolor=black,citecolor=blue}
\color{black}
A 200-sample preview subset of A2A-UAV is hosted on Google Drive\cite{a2auav_data} at \url{https://drive.google.com/file/d/1oCtjoF5vfH3MsPDNQEOQ84MvRywHSUso/view}, and an identical mirror is provided on Baidu Netdisk at \url{https://pan.baidu.com/s/1_P6F3z18gCyuDagdedcM2Q?pwd=A2AD} with the extraction code A2AD.
\endgroup
The full dataset includes multimodal sensor products, sample-level metadata, RGB and panoramic bounding-box annotations, and predefined training, validation, and test splits. The data supporting the technical validation are included in the full dataset, which will be publicly released upon acceptance of this paper. The preview subset provides the same data products for 200 samples. Both the preview subset and the full dataset are distributed under the Creative Commons Attribution 4.0 International (CC BY 4.0) license.

\section*{Code Availability}

\begingroup
\hypersetup{urlcolor=black}
\textcolor{black}{The custom code for data processing and benchmark evaluation is available at \textcolor{black}{\url{https://github.com/lht-research/A2A-UAV}}. The code release includes data-loading examples, annotation processing, panoramic tiling and prediction merging, and training and evaluation configurations for the detection and identification experiments. Software dependencies and reproduction instructions are provided in the accompanying documentation.}
\endgroup

\section*{Author Contributions}

\textcolor{black}{H.L. and Y.W. contributed equally to this work. H.L. conceived the dataset, designed the airborne sensing platform, led the data acquisition, developed the data processing pipeline, and contributed to the manuscript writing. Y.W. designed the benchmark experiments, performed the technical validation, and wrote the manuscript. K.Z., J.Z., Z.J., and Z.Z. participated in the flight experiments and the data acquisition, and performed the data annotation and quality inspection. J.J. contributed to the benchmark evaluation and the analysis of the validation results. F.Z. supervised the project, provided the funding, and revised the manuscript. Q.W. supervised the project and provided the research resources. All authors reviewed and approved the final manuscript.}

\section*{Competing Interests}

The authors declare no competing interests.

\section*{Funding}

\textcolor{black}{This work was supported in part by the Interdisciplinary Innovation Fund for Doctoral Students of Nanjing University of Aeronautics and Astronautics under Grant KXKCXJJ202607, and in part by the Postgraduate Research and Practice Innovation Program of Jiangsu Province under Grant 26CXJH0870.}

\begingroup
\color{black}
\hypersetup{urlcolor=black}
\renewcommand{\bibnumfmt}[1]{#1.}

\endgroup

\end{document}